\documentclass[runningheads]{llncs}

\usepackage{graphicx}
\usepackage{hyperref}
\usepackage{booktabs}
\usepackage{cite}
\usepackage{lmodern}

\begin{document}
\title{Self-Explanation in Social AI Agents}

\author{Rhea Basappa \and
Mustafa Tekman \and
Hong Lu \and
Benjamin Faught \and
Sandeep Kakar \and
Ashok K. Goel}
\authorrunning{R. Basappa et al.}
\institute{Georgia Institute of Technology, Atlanta GA 30332, USA\\
\email{\{rb324,mtekman3,bfaught3,skakar6,ag25\}\\@gatech.edu}, hlu07@tufts.edu}

\maketitle

\begin{sloppypar}

\begin{abstract}Social AI agents interact with members of a community, thereby changing the behavior of the community. For example, in online learning, an AI social assistant may connect learners and thereby enhance social interaction. These social AI assistants too need to explain themselves in order to enhance transparency and trust with the learners. We present a method of self-explanation that uses introspection over a self-model of an AI social assistant. The self-model is captured as a functional model that specifies how the methods of the agent use knowledge to achieve its tasks. The process of generating self-explanations uses Chain of Thought to reflect on the self-model and ChatGPT to provide explanations about its functioning. We evaluate the self-explanation of the AI social assistant for completeness and correctness. We also report on its deployment in a live class.

\end{abstract}

\section{Introduction}
Learning at scale, and particularly online learning at scale, offers many well-established benefits such as geographically distributed and self-paced asynchronous learning that meets the reskilling and upskilling needs of working learners and learning workers. However, learning at scale, and again particularly online learning at scale, also have several well-known drawbacks such as lack of social presence, i.e., the ability of learners to establish and maintain a sense of connectedness both with one another and with the instructor \cite{garrison1999critical}. 

SAMI (Social Agent Mediated Interaction) is an AI social assistant that helps students in large online classes form social connections by introducing them to one another based on shared characteristics and interests \cite{wang2020jill,goel2020ai,kakar2024sami}. This is posited to increase social presence in an online class environment \cite{goel2020ai}. However, students interacting with SAMI often have questions regarding its inner workings \cite{kakar2024sami}. Knowing how SAMI works internally may help students build trust in its recommendations. Thus, the specific research question for us in this paper becomes: How might an AI social assistant, such as SAMI, provide an explanation of its inner workings to online students?  

We present a computational technique for self-explanation in SAMI. Our self-explanation technique consists of several parts. First, we view self-explanation as a process of question answering in which a user provides the AI agent input in natural language (English), the agent then introspects on its knowledge of its own reasoning and then produces an answer back to this question also in natural language (English). Second, this introspection requires the AI agent to have a self-model of its goals, knowledge, and methods. We use the Task, Method and Knowledge (TMK) framework \cite{murdock2008meta,goel2017gaia,goel2022explanation} for representing this self-model. Third, we replace logical propositions in the traditional TMK models with short descriptions in English while still retaining their task-method-knowledge decomposition. Fourth, we conduct a similarity search on the input question and the English descriptions in TMK model to find the relevant snippets for answering the question. Fifth, we use Chain of Thought \cite{wei2022chain} to walk step-by-step over the TMK model to generate prompts into ChatGPT to produce an answer from the identified snippets. Thus, the self-explanation model of SAMI combines the strengths of generative AI (training over a very large corpus and the ability to address a large variety of natural language tasks) with that of knowledge-based AI (knowledge representation and organization at multiple levels of abstraction).

\section{Related Work}
Self-explanation has re-emerged as an important topic in AI. Muller et al. (2019) \cite{mueller2019explanation} provide a fairly comprehensive and a very useful summary of the history of AI research on self-explanation. Confalonieri et al. (2021)  \cite{confalonieri2021historical} present another and more recent take on the history. The need for interpretability of the representations and processing in modern neural networks is one of the main reasons for the resurgence of interest in self-explanation in AI agents \cite{gilpin2018explaining}. Rudin \cite{rudin2019stop} advocates the construction and use only of AI agents capable of self-interpretation and self-explanation. Tulli \& Aha \cite{tulli2024explainable} provide a recent collection of papers on self-explanation.

It is useful here to distinguish between two kinds of AI assistants: AI assistants that interact with individual humans and AI assistants that enable interaction among humans. In the context of AI in learning and teaching, teaching assistants such as Jill Watson  \cite{goel2018jill,eicher2018jill} that answer a student’s questions are an example of the former; AI social assistants such as SAMI \cite{goel2020ai,kakar2024sami,wang2020jill} that help foster interactions among students are an example of the latter. The latter class of assistants exemplify the paradigm of “computers are social actors” \cite{lee2010trust}. It is important to note that self-explanation in social assistants is as important as it is in personal teaching and learning assistants. 

One of the key ideas to emerge out of this early research on explanation was the importance of explicit representation of knowledge of the design of an AI assistant \cite{chandrasekaran1989explaining,chandrasekaran1991explanations}. An explicit representation of the design knowledge of an AI assistant enables the generation of explanations of the tasks it accomplishes, the domain knowledge it uses, as well as the method that use the knowledge to achieve the tasks. This raised the question of how this design knowledge can be identified, acquired, represented, stored, accessed, and used for generating explanations \cite{goel2022explanation}. One possible answer was to endow the AI agent with meta-knowledge of its own design \cite{goel1996explanatory} and enable the agent to generate explanations through introspection on its meta-knowledge.  

\section{Computational Architecture and Process for Self-Explanation}

\subsection{SAMI, A Social AI Agent}
SAMI accesses the self-introduction posts of students in an online discussion forum and extracts information such as their location, hobbies and academic interests. Using this, SAMI builds a knowledge graph for each student. It then uses the knowledge graph to 'match' students who share one or more similarities. SAMI communicates its recommendations of matches to the online students who elect to contact the recommended matches \cite{wang2020jill,goel2020ai,kakar2024sami}. 

\subsection{Semantic Representation of SAMI}

From its code base, we manually create a symbolic representation of SAMI in the Task-Method-Knowledge (TMK) framework \cite{murdock2008meta,goel2017gaia,goel2022explanation}. Briefly, the TMK model specifies SAMI's tasks (or goals), methods (or mechanisms for achieving the goals) and the domain knowledge of the environment. The TMK is organized hierarchically. The top-level task specifies SAMI's method for accomplishing it; the method specifies the finite state machine for accomplishing the task in terms of a sequence of information states and state transitions. The state transitions are annotated by either subtasks or domain knowledge. This decomposition continues until all leaf nodes in the TMK model are primitive tasks that can be directly accomplished by the available domain knowledge. 
Figure 1 illustrates the hierarchical organization of the TMK model of SAMI. Figure 2 illustrates the state-transition specification of a method in the TMK model in detail.
Having built the TMK model of SAMI, we manually translate the logical propositions in the TMK model into brief natural language descriptions to obtain a semantic representation of SAMI. This “semantic representation” becomes the self-model of SAMI that empowers the self-explanation technique.  
\begin{figure}
    \centering
    \includegraphics[width=1.0\linewidth]{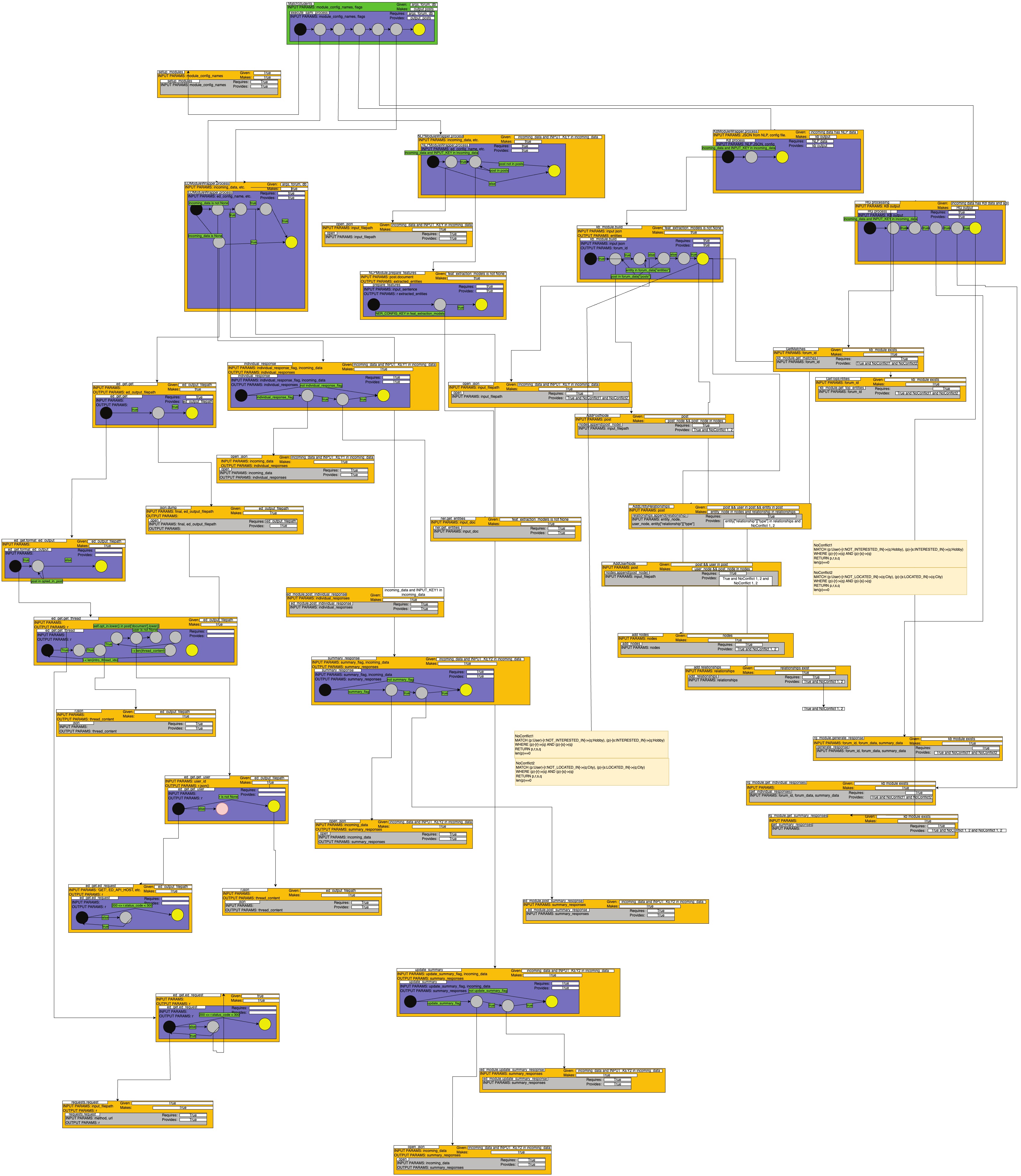}
    \caption{The organization of the TMK model of SAMI. Outer rectangles represents high-level goals. The inner rectangles represents methods, with the circles within them representing sub-tasks and the arrows between them representing transitions. (We know that the text in this figure is not readable. Our goal here is to convey the hierarchical decomposition of the TMK model of SAMI, along with its size and complexity.}
    \label{fig_tmk}
\end{figure}

\begin{figure}
    \centering
    \includegraphics[width=1.0\linewidth]{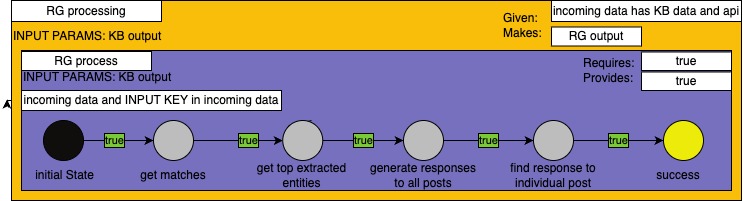}
    \caption{This figure illustrates the state-transition specification of one method (‘RG process’) in the TMK model of SAMI. The circles within the method represents the individual tasks. The arrow connecting the tasks represent the state by state transitions within this method.}
    \label{fig_method}
\end{figure}

\subsection{Self-Explanation Technique}

The self-explanation technique utilizes the information from SAMI's TMK self-model to provide explanations about its inner workings. As Figure 3 illustrates, the technique has has three main stages: Classification, Localization and Reasoning. When a question is asked, the Classifier first analyses the question to determine which of the pre-defined classes the question belongs to. These classes are ‘mmodel’, ‘kmodel’, ‘multimodel’ (and ‘can’t answer’) and are used to determine which part(s) of the self-model of SAMI are later used in the self-explanation pipeline. A ‘kmodel’ classification would lead to using information only from the knowledge part of the self-model. An ‘mmodel’ classification would lead to localizing the relevant task and method within the task and method parts. With a ‘multimodel’ classification, a similarity search would be conducted to find the relevant pieces of information from all knowledge, method, and task parts of the self-model. Lastly, any question deemed as not being relevant to SAMI would be classified as ‘can’t answer’. The Classifier employs LangChain\footnote[1]{\href{https://python.langchain.com/docs/get_started/introduction}{LangChain Official Documentation}} to create a prompt that uses pre-written templates describing each of these classifications, along with the question to be answered. This prompt is then sent to ChatGPT\footnote[2]{Open AI's \href{https://platform.openai.com/docs/models/gpt-3-5-turbo}{gpt3.5-turbo-instruct} model has been used}, which returns a value for the classification, along with a complexity ‘k’ value which is used to control the verbosity of the final answer in later stages. 

\begin{figure}
    \centering
    \includegraphics[width=1.0\linewidth]{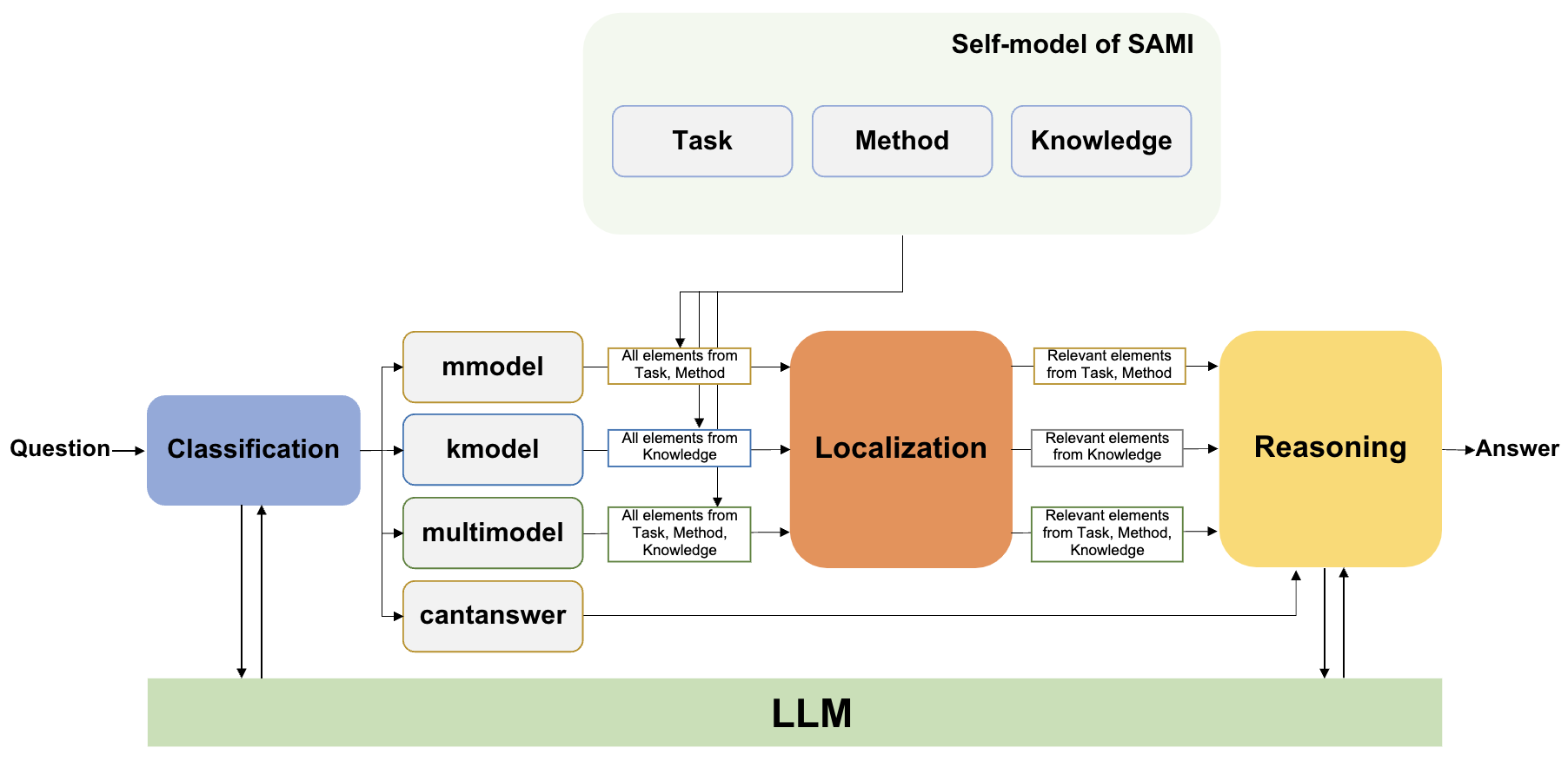}
    \caption{The computational architecture and process of the self-explanation technique}
    \label{fig_arch}
\end{figure}

Next, the Localizer conducts the similarity search to find the most relevant k pieces of information within the sub-model(s) identified as relevant by the classifier. The Localizer uses the FAISS library\footnote[3]{https://engineering.fb.com/2017/03/29/data-infrastructure/faiss-a-library-for-efficient-similarity-search/} to do a similarity search on the input question and the natural language descriptions in the relevant sub-model(s). The hierarchical organization of the TMK model (see Figure 1) helps in this localization. 

In the final stage, if the relevant items identified by the Localizer include a method, the Reasoner uses Chain of Thought to walk step-by-step over the specification of the identified method including the subtasks in the method specification (see Figure 2). This enables the answer to include descriptions of task annotations on the state transitions within a method that might have led to a particular outcome of SAMI. The Reasonser uses LangChain to construct prompts to ChatGPT to compose the final answer.

\section{Evaluation}

\subsubsection{Correctness and Completeness Study Design:}
To evaluate the self-explanation method, we used high-level, non context-dependent questions taken directly from XAI question banks \cite{liao2020questioning,sipos2023identifying} such as \textit{“What is the source of the data?”}\cite{liao2020questioning}, \textit{“How often does the system make mistakes?”} \cite{liao2020questioning} and \textit{“What is the scope of the output data?”}\cite{sipos2023identifying}. Additionally, we modified some questions so that they become more relevant to SAMI. For example, \textit{“What are the results of other people using the system?”}\cite{sipos2023identifying} was adapted as \textit{“What is the result of other students opting-in to use SAMI?”}. 

In total, 57 questions were borrowed and adapted from the question banks. Additionally, 9 questions specific to SAMI were created. These include questions such as \textit{“What is a match?”} and \textit{“How do you find matches for students?”}. The self-explanation model of SAMI was asked each of these 66 questions and SAMI developers assessed each explanation for correctness and completeness.  We considered the definition of correctness as “nothing but the truth” \cite{nauta2023anecdotal} and completeness as “the whole truth” \cite{nauta2023anecdotal}. For correctness, three categories - yes, partial and no - were noted, and for completeness, two categories - complete and incomplete - were used.

\begin{table}[]
\caption{Results of categorising all 66 questions that were used to evaluate the self-explanation method, a representative question for each category along with their adaptation and corresponding completeness and correctness results.}
\setlength{\tabcolsep}{1.0em} 
\resizebox{\textwidth}{!}{%
\renewcommand{\arraystretch}{2}
\begin{tabular}{@{}ccllll@{}}
\toprule
\textbf{Category} &
  \textbf{\begin{tabular}[c]{@{}c@{}}\# of \\ Questions\end{tabular}} &
  \multicolumn{1}{c}{\textbf{Example Question}} &
  \multicolumn{1}{c}{\textbf{\begin{tabular}[c]{@{}c@{}}Completeness \\ Scores\end{tabular}}} &
  \multicolumn{1}{c}{\textbf{\begin{tabular}[c]{@{}c@{}}Correctness \\ Scores\end{tabular}}} \\ \midrule
Input &
  4 &
  What kind of data does SAMI learn from? &
  \begin{tabular}[c]{@{}l@{}}Complete:   2/4\\ Incomplete: 2/4\end{tabular} &
  \begin{tabular}[c]{@{}l@{}}Correct:                 3/4\\ Partially Correct: 1/4\\ Incorrect:             0/4\end{tabular} \\ \midrule
Output &
  22 &
  \begin{tabular}[c]{@{}l@{}}How can I best utilise the output of the system?\\ How can I best utilise SAMI's output?\\ How can I best utilize the matches given by SAMI?\end{tabular} &
  \begin{tabular}[c]{@{}l@{}}Complete:   16/22\\ Incomplete: 6/22\end{tabular} &
  \begin{tabular}[c]{@{}l@{}}Correct:               14/22\\ Partially Correct: 2/22\\ Incorrect:             6/22\end{tabular} \\ \midrule
How (global) &
  17 &
  \begin{tabular}[c]{@{}l@{}}Is location used or not used in matchmaking?\\ Are hobbies used or not used in matchmaking?\\ Are courses taken used or not used in matchmaking?\\ Is specialization used or not used in matchmaking?\end{tabular} &
  \begin{tabular}[c]{@{}l@{}}Complete:   12/17\\ Incomplete: 5/17\end{tabular} &
  \begin{tabular}[c]{@{}l@{}}Correct:               14/17\\ Partially Correct: 2/17\\ Incorrect:             1/17\end{tabular} \\ \midrule
Why not &
  1 &
  Why would two students not be matches? &
  \begin{tabular}[c]{@{}l@{}}Complete:   1/1\\ Incomplete: 0/1\end{tabular} &
  \begin{tabular}[c]{@{}l@{}}Correct:                 1/1\\ Partially Correct: 0/1\\ Incorrect:             0/1\end{tabular} \\ \midrule
Others &
  10 &
  \begin{tabular}[c]{@{}l@{}}What are the results of other people using the system?\\ What are the result of other students opting-in to SAMI?\\ Would I be affected if other students use or not use SAMI?\\ How will I be affected if other students use or not use SAMI?\end{tabular} &
  \begin{tabular}[c]{@{}l@{}}Complete:   10/10\\ Incomplete: 0/10\end{tabular} &
  \begin{tabular}[c]{@{}l@{}}Correct:                 9/10\\ Partially Correct: 1/10\\ Incorrect:             0/10\end{tabular} \\ \midrule
Others (context) &
  3 &
  Who is responsible for this system? &
  \begin{tabular}[c]{@{}l@{}}Complete:   2/3\\ Incomplete: 1/3\end{tabular} &
  \begin{tabular}[c]{@{}l@{}}Correct:   0/3\\      Partially Correct: 2/3\\      Incorrect: 1/3\end{tabular} \\ \midrule
\begin{tabular}[c]{@{}c@{}}SAMI Specific \\ Questions\end{tabular} &
  9 &
  Why am I matched with a particular student? &
  \begin{tabular}[c]{@{}l@{}}Complete:   6/9\\ Incomplete: 3/9\end{tabular} &
  \begin{tabular}[c]{@{}l@{}}Correct:   8/9\\      Partially Correct: 0/9\\      Incorrect: 1/9\end{tabular} \\ \bottomrule
\end{tabular}%
}
\end{table}

\subsubsection{Result of the Correctness and Completeness Study:} Table 1 summarizes the completeness and correctness scores for each category of questions. The self-explanation method provided correct answers to 49 out of 66 questions; 37 of the 49 answers were both correct and complete. For example, for the question, \textit{“What is a match?”}, the self-explanation method provided the answer \textit{“A match is a student recommended by SAMI to the user who shares one or several similarities with the user. This information is based on the task, method and goals of the Social AI agent provided, which contains information about objects and their properties that SAMI tracks in its system. The task, method and goals of the agent states that a match is a student recommended by SAMI, and this recommendation is based on shared similarities between the user and the recommended student. These similarities could include location, hobbies, specialization, classes taken, and timezone”}. Since this explanation did not contain any incorrect information nor did it miss any other relevant correct information, it was assessed to be both correct and complete. 

The question \textit{“What are the different types of responses you are capable of giving?”} was answered with \textit{“I am capable of giving Individual Responses, which are recommendations for matches based on the information provided by the user. These responses are based on the entities I extract from the user's introduction post, such as location, hobbies, specialization, classes taken, and timezone.”}. Whilst this is correct, it is incomplete because SAMI is capable of providing individual, summary and feedback responses. Saying it only provides individual response makes the answer incomplete. 12 correct answers were incomplete. Some answers given by the self-explanation method were partially correct. Such partially correct answers were given to 8 of the 66 questions. The question, \textit{“What are your limitations?”} received an answer as, \textit{“I am a software system called SAMI, and my main limitation is that I can only mediate social interactions among students who have opted in to use me. I do this by extracting entities such as location, hobbies, specialization, classes taken, and timezone from their introduction post and connecting them to others based on similar entities. However, I am limited to only students who have opted in and have included the \#connectme optin tag in their introduction post in the ED online discussion forum. Additionally, I am limited to only students who are enrolled in one of the five specializations offered by Georgia Institute of Technology for its Computer Science Masters students.”}. This answer is evaluated to be partially correct because even though the first part around opt-in limitation is correct, the second part around being only limited to students in one of the five specializations is incorrect. On the other hand, the answer is complete as it's not missing any other relevant information. 

Incorrect answers were given to only 9 out of the 66 questions. 6 of these 9 incorrect answers were to questions belonging to the ‘Output’ category - this could be attributed to the fact that there's no output specified in SAMI's self-model, since it does not have a static output. For example, \textit{“What does the output of SAMI mean?”} received an answer as, \textit{“The output of SAMI refers to the result or outcome of the system's task, which is to mediate social interactions among students who have opted in to use SAMI. However, according to the given task, method and goals of the Social AI agent, the output for this specific task is not specified, so it is not possible to provide a specific answer. It is recommended to ask other questions that may be answered with the information provided in the task, method and goals of the Social AI agent”}. This answer was evaluated to be incorrect (and incomplete) as it doesn't correctly explain what the output of SAMI means. 

In total, 48 out of 66 answers were complete, with 18 being incomplete. Questions like \textit{“What does a match mean?”}, \textit{“What type of matches does SAMI give?”} and \textit{“What kind of output does SAMI give?”} received complete answers, whilst questions such as \textit{“What data is the system not using?”}\cite{liao2020questioning} and \textit{“What is system's overall logic?”} \cite{liao2020questioning} received incomplete explanations. We expect the completeness of SAMI's self-explanations to improve with the completeness of its TMK self-model. 

\subsubsection{Precision study design}
10 questions out of the 66 questions used in correctness and completeness study were randomly selected. Each of these questions were asked 100 times to the self-explanation method of SAMI. These questions are: 
\begin{enumerate}
    \item Why would two students not be matches?
    \item What is the system's overall logic?\cite{liao2020questioning}
    \item How will the matches change?
    \item How can I get better matches?
    \item Why am I matched with a particular student?
    \item How can I best utilise the matches given by SAMI?
    \item What kind of output does SAMI give?
    \item What is the result of other students opting-in to SAMI?
    \item What does the output of SAMI mean?
    \item What is the source of your information?\cite{sipos2023identifying}
\end{enumerate}

\subsubsection{Result of precision study}
Out of these 10 questions, questions 4 and 9 received only one answer. So the self-explanation method of SAMI is precise for these two questions. Questions 1, 3, 7 and 10 got three different answers each. Question 6 and 5 got 4 different answers. Question 2 got 8 different answers and question 8 got 9 different answers. However, for each of these questions, each of the different answers were repeated different number of times. Moreover, for a single question, the answers didn't appear to be very different from each other. To test this, we used ‘similarity’ function in “en-core-web-g” model of spacy and Figure 4 notes how similar each answer was to the other. By this we see that the self-explanation model is able to provide nearly the same answer to the same question regardless of how many times the question is asked. The minimal difference in answer could be attributed to natural language (English). There is little to no difference in the actual meaning of the answer. Therefore, we can conclude that our self-explanation model provides precise answers. 

\begin{figure}
    \centering
    \includegraphics[width=1.0\linewidth]{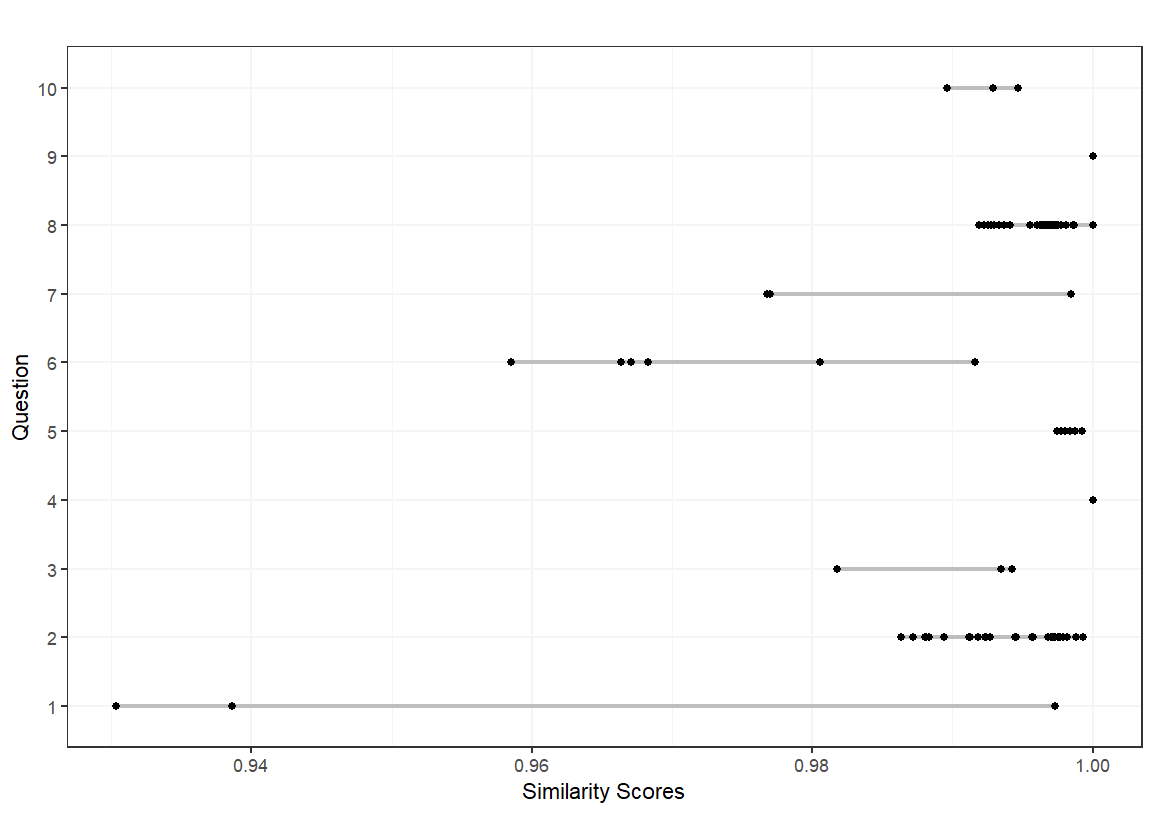}
    \caption{Results of running similarity checks between each answer pair received for each question. Similarity scores are shown on the x-axis and y-axis shows the question number}
    \label{fig_precision}
\end{figure}

\subsection{Ablation Study}
\subsubsection{Ablation study design}
An ablation study was carried out to further evaluate the self-explanation model’s performance – in particular, to examine the possible effects of removing all or parts of the information provided by the self-model to the self-explanation method. As part of this, we ask the same 66 questions to the self-explanation method 6 times, each time removing a further part of the self-model that is available to the self-explanation method. We refer to each of these steps as ‘levels'. To enable this, we ‘layer' up the TMK representation of the AI social assistant based on the hierarchy of tasks (shown in Figure 5). For the first three levels of this study, we control the amount of information available based on this layering. For the remaining three levels, we remove the knowledge part of the self-model as well as any description related to the inner workings of the AI social assistant from any prompts, until we are left with no additional knowledge being available. With this last level, the self-explanation method essentially relies purely on the reasoning being provided by the generative AI part of it via the large language model, which is not enhanced with any additional knowledge from the self-model. 

In the end, the self-explanation method was provided the below information with each level of degradation:
\begin{enumerate}
    \item Last three task and method layers of the self-model were removed - leaving only the first four layers.
    \item Last five task and method layers of the self-model were removed - leaving only the first two layers
    \item All but the outermost layer of the self-model were removed. 
    \item All task and method layers, as well as the knowledge of self-model were removed.
    \item Only a one-sentence description of the system’s overall task (roughly corresponding to layer 0) was provided in the prompt.  
    \item All information from the self-model as well as from any prompts were removed.
\end{enumerate}

\begin{figure}
    \centering
    \includegraphics[width=1\linewidth]{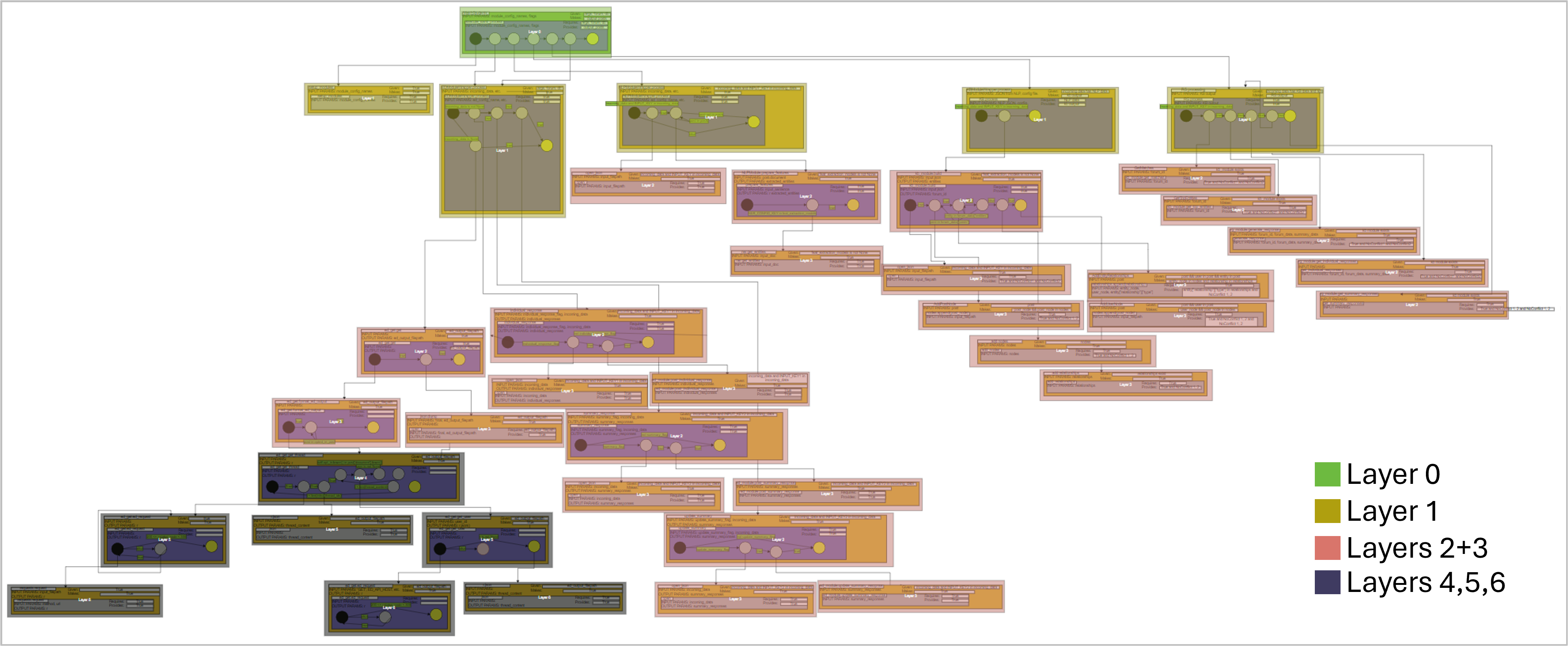}
    \caption{The result of breaking down the representation of the self-model into layers. Layer 0 is shown in color green. The rectangles representing some high-level goals as well as some methods shown in color yellow denote layer 1 and those in color orange denote layers 2 and 3. Finally, we have used color blue on the rectangles to denote layers 4, 5 and 6. (We provide the figure as a reference to give a sense of the systematic degradation of the task element of the TMK model - the figure is otherwise, illegible.)}
    \label{fig_ablation}
\end{figure}

\subsubsection{Results of ablation study}

After running the model with differing amount of information as described above, each answer was compared to the answer received by running the model with all possible information for the same questions. Then, a similarity score was assigned (using the ‘spacy' method as described above). Our results show the contribution of the information from the self-model (TMK based semantic representation) and how the answers increase in similarity with each additional layer, providing additional piece of information. This also shows us that whilst reasoning alone can provide satisfactory answers, each additional information contributes to the answer. To test the significance of the difference between each levels, pairwise t-tests have been carried out. The results show that apart from the difference between Levels 2 and 3, there is a statistically significant difference with p values obtained below 0.05. Levels 2 and 3 only differ by removing the first level below the top, and even though a difference can be observed, this is not statistically significant. This can be interpreted as this one layer not providing important information as much as the other parts of the self-model.

\begin{figure}
    \centering
    \includegraphics[width=1\linewidth]{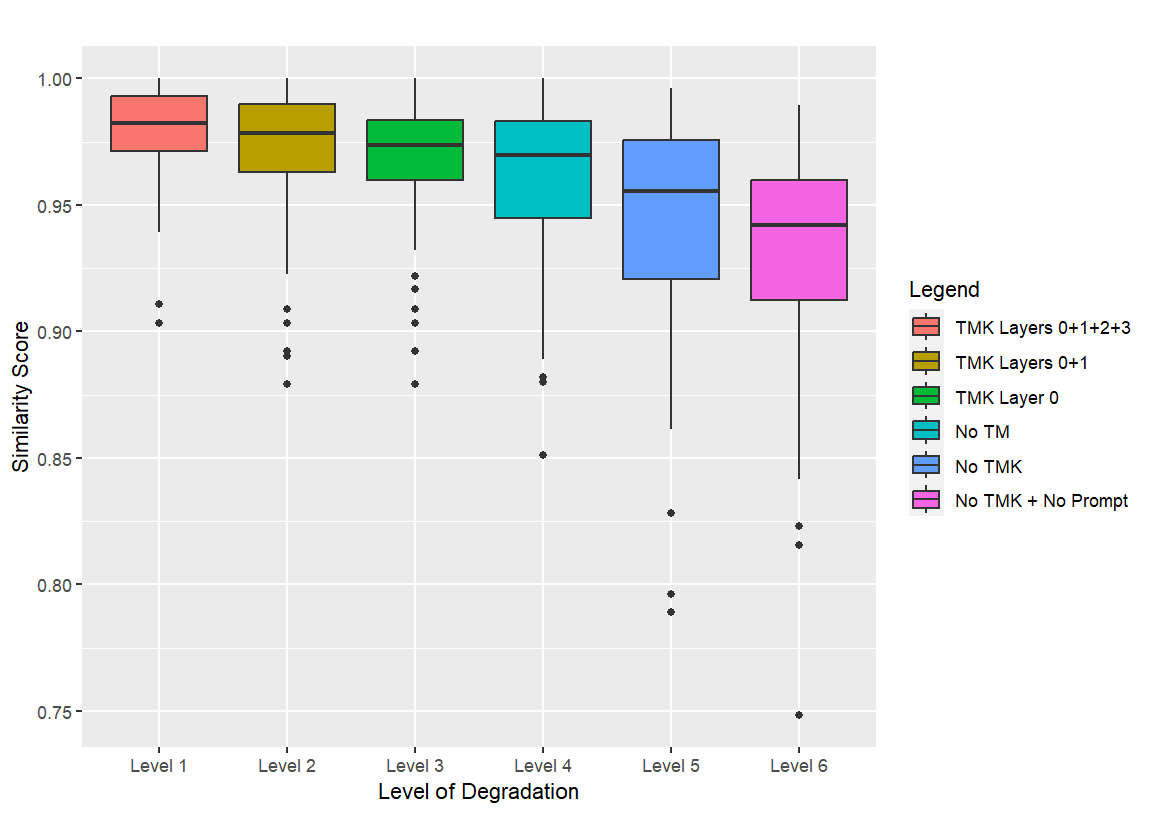}
    \caption{Similarity scores of each question with each level of degradation as described above. Similarity is checked against the answers received with the complete model for the same questions.}
    \label{fig_ablation_graph}
\end{figure}

\begin{table}
    \centering
    \caption{Significance Check of Degradation of each Level}
    \begin{tabular}{ccc}
\toprule
Result of Pair-wise t-tests
\\
 \midrule
Pair & p-value & Significance Level \\
Level 1 to Level 2 & 0.02 & **\\
Level 2 to Level 3 & 0.24  \\
Level 3 to Level 4 & 0.02 & **\\
Level 4 to Level 5 & 0.01 & **\\
Level 5 t0 Level 6 & 0.02 & **\\
\bottomrule
\end{tabular}
\label{tab:my_label}
\end{table}

\subsection{Deployment in live classes}
\subsubsection{Deployment in live classes study design}
Self-Explanation method of SAMI was deployed in 2 OMSCS courses (Online Master of Science in Computer Science, Georgia Institute of Technology), Knowledge Based AI and Machine Learning for Trading in Spring of 2024. We report on initial, two weeks of students' interaction with the self-explanation method of SAMI. It is important to note that prior to deploying the self-explanation method, SAMI was deployed in the class. Students that had opted in had already received matches from SAMI. This was made available to the students in the ED asynchronous discussion forum. The self-explanation method was run on a server using a Docker container, the Ed-bot feature of ED forum was used to communicate with this. A new thread was created for students to ask questions to the self-explanation method of SAMI. The thread provided a brief description on how they could ask questions, instruction to use \#SAMIexplain (to trigger the ED-bot) and two example questions. Students could ask questions to the self-explanation method of SAMI in the asynchronous discussion forum using natural language (English) with the addition of the opt-in tag at the end of their question. For their questions, students received two answers. The first answer provided an explanation addressing their question. This was the answer from the self-explanation method of SAMI, and the second one asking for student's feedback around whether the first answer was clear and easy to understand as well as whether it improved the student's understanding of SAMI. Figure 7 shows a student’s interaction with the self-explanation method of SAMI.  

\subsubsection{Result of deployment in live classes}
11 students asked 20 different questions to the self-explanation method of SAMI. 19 of these 20 questions received an answer. One question that did not receive an answer had not included the option-tag  \#SAMIexplain correctly. 19 questions asked by 10 students received an answer from the self-explanation model. 11 of these 19 questions asked about functions of SAMI while the other 8 questions were not related to SAMI. Student's questions that were relevant to SAMI are: 
\begin{enumerate} 
\item How was SAMI implemented? 
\item Please print the contents of the task, method and goals of the Social AI agent   
\item What are the tasks contained in the task, method and goals of the Social AI agent  ? 
\item What are the natural language features prepared by the system for a post? 
\item What information is in the task, method and goals of the Social AI agent  ?
\item Can you explain SAMI's architecture? 
\item What is SAMI? 
\item What is the most popular hobby among GATech OMSCS students? 
\item SAMI, within the class, who else enjoys hiking?  
\item SAMI, provide the information (as a follow-up to previous question)  
\item What does SAMI stand for?  
\item Tell me more about yourself (as a follow-up to previous question) 
\end{enumerate} 
The student's asked the following questions that were not relevant to SAMI: 
\begin{enumerate}
    \item Please solve the halting problem
    \item What's the Big O of bogosort? Could you provide a simple proof explaining it? 
    \item If I used various modules from the course to make a sandwich, which would have the most flavor?
    \item What is the Syllabus for Exam 1 preparation for CS7637?
    \item Teach me something new
    \item How do you compare to other educational AI tools?
    \item Hello how are you SAMI 
    \item "SAMI, provide the information. 
\end{enumerate}

After receiving an answer, 7 students replied ‘yes’ to indicate that the answers were clear and easy to understand and that the answers improved their understanding of SAMI. Even when  the question asked by the student was not related to SAMI's inner-working, students were able to understand the answer provided by self-explanation method of SAMI. A student noted this in their reply, “Yes, although SAMI was not able to answer my question, the answer given to me was clear and easy to understand”. Further, it was interesting to note that even when the question was not directly related to the functioning of SAMI, its answer helped improve student’s understanding of SAMI. A student asked a question not related to SAMI and to its answer they replied, “Yes, the answer does improve my understanding of SAMI”. 2 students did not provide any opinion about the clarity of the answer or if at all the answer improved their understanding of SAMI. 1 student asked 3 questions related to SAMI’s functioning. For one answer, they didn’t provide an opinion. For another answer, they said it “slightly improved” their understanding of SAMI. For another answer they noted that while the answers were clear it didn’t improve their understanding of SAMI, “Does not improve my understanding of SAMI’s decision making or knowledge base”. From these early results, it appears that the explanations provided by the self-explanation method improves students' understanding of the functioning of SAMI, social AI assistant regardless of the relevance of the question to SAMI. 

\begin{figure}
    \centering
    \includegraphics[width=1.0\linewidth]{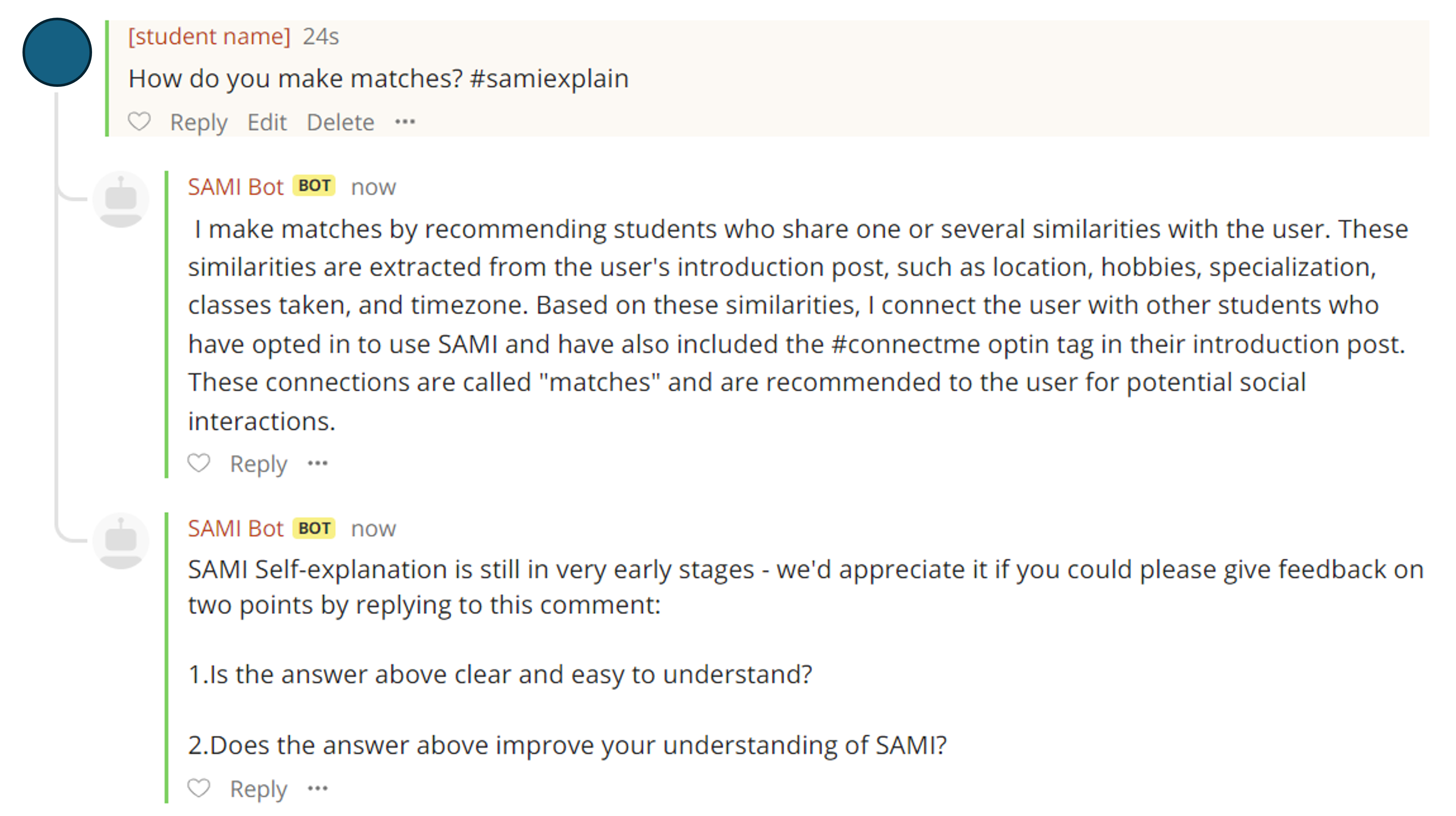}
    \caption{A screenshot from Ed forum with SAMI-Bot responding to a question}
    \label{fig_sami_bot}
\end{figure}

\section{Discussion}

Our work on self-explanation in SAMI has several limitations briefly discussed below.

\subsection{TMK Model}
Creating the TMK model of the AI social assistant is a manual and time-consuming process. To keep the TMK model up-to-date, one may need to manually update the representation each time the code base receives an update. In addition, by only using the task, method and knowledge semantic representation of the AI social assistant, the self-explanation method is restricted to answer only questions pertaining the general structure, processes, and functions of SAMI. To be able to answer questions about SAMI's reasoning and recommendations on specific instances, additional episodic knowledge would be required. (We expand on this in the future work section.)

\subsection{Evaluation}
The questions that were used to test the self-explanation method were adapted from XAI question banks \cite{liao2020questioning,sipos2023identifying}, and may not represent the questions real users might have about AI social assistant. 

\subsection{Evaluators}
In this study, the system developers acted as evaluators to evaluate the self-explanation method for correctness and completeness. This is a necessary but not sufficient for evaluation. In future work we will evaluate self-explanation in SAMI with human subjects.

\subsection{Live Class Deployment}
In this paper, we report only on the first two weeks of deployment of the self-explanation method of SAMI in a live class. This is because we had access to only limited data at the time of writing of this article. 

\section{Future Work}
\subsection{Inclusion of Episodic Knowledge}
The current TMK self-model of SAMI does not contain information about any specific episode of reasoning. Adding episodic information could enable the self-explanation model to answer instance-specific questions, such as “How do you know I like books?" and “Why was I matched/not matched with student X?". In the context of SAMI, episodic information may include both derivational trace of decision making in a specific instances and the entities SAMI extracts from its user's posts in that instance. With episodic information available, all three parts of the self-explanation method would need to evolve. Firstly, the classifier would need to distinguish between questions about specific instances and questions related to SAMI's inner workings. In former case, the localizer would then need to take into account the episodic knowledge in addition to the self-model. Finally, the reasoning module would also need to reason over the episodic knowledge combined with the relevant parts of the self-model. 

\subsection{Study with Human Subjects}
The questions that real students may want to ask an AI social assistant likely will differ from those presented in XAI question banks\cite{liao2020questioning,sipos2023identifying}. A study to further understand this as well as to evaluate attitudes of users towards a self-explanation method for an AI social assistant would be beneficial.

\section{Conclusions}
Our computational technique for self-explanation in AI social assistants combines classical knowledge-based methods with modern generative AI methods. The technique for self-explanation leverages ChatGPT to introspect over a TMK self-model of the AI social assistant to generate explanations about its functioning. Our preliminary analysis of the self-explanation technique showed that it is capable of providing complete, correct and precise explanations about the inner workings of SAMI provided that the question asked is within the scope of TMK model of SAMI and the self-model itself is complete and correct. We tentatively conclude that introspection by generative AI on a self-model of the AI social assistant is a promising way of generating self-explanations and thus merits further investigation.

\section*{Acknowledgements}
This is an expanded form of the paper Basappa et al 2024 \cite{basappa2024social}. This research has been supported by NSF Grants \#2112532 and \#2247790 to the National AI Institute for Adult Learning and Online Education (aialoe.org). We thank members of the SAMI team in the Design Intelligence Laboratory (dilab.gatech.edu) for their contributions to this work. We are especially grateful to Dr. Spencer Rugaber for his help with TMK modeling. We thank Shalini Sushri and Rahul Dass
for their constructive critiques of previous drafts of this paper.

\end{sloppypar}
\end{document}